\documentclass[sigplan]{acmart}
\renewcommand\footnotetextcopyrightpermission[1]{} 
\usepackage{tabularx}
\usepackage{subcaption}

\AtBeginDocument{%
  \providecommand\BibTeX{{%
    \normalfont B\kern-0.5em{\scshape i\kern-0.25em b}\kern-0.8em\TeX}}}

\setcopyright{acmcopyright}
\copyrightyear{2018}
\acmYear{2020}
\acmDOI{10.1145/1122445.1122456}

\acmConference{KDD Workshop on ML in Finance 2020}{August 24, 2020}{San Diego, CA, USA}

\begin{document}

\title{Accurate and Intuitive Contextual Explanations using Linear Model Trees}


\author{Aditya Lahiri}
\affiliation{%
  \institution{American Express, AI Labs}
    \city{Bangalore}
 \state{Karnataka}
 \country{India}
  }
\email{Aditya.Lahiri1@aexp.com}

\author{Narayanan Unny Edakunni}
\affiliation{%
  \institution{American Express, AI Labs}
  \city{Bangalore}
 \state{Karnataka}
 \country{India}
  }
\email{Narayanan.U.Edakunni@aexp.com}

\begin{abstract}
With the ever-increasing use of complex machine learning models in critical applications within the finance domain, explaining the model’s decisions has become a necessity. With applications spanning from credit scoring\cite{b8} to credit marketing\cite{marketing}, the impact of these models is undeniable. Among the multiple ways in which one can explain the decisions of these complicated models\cite{b12}, local post hoc model agnostic explanations have gained massive adoption. These methods allow one to explain each prediction independent of the modelling technique that was used while training. As explanations, they either give individual feature attributions\cite{b1} or provide sufficient rules that represent conditions for a prediction to be made\cite{b2}. The current state of the art methods use rudimentary methods to generate synthetic data around the point to be explained. This is followed by fitting simple linear models as surrogates to obtain a local interpretation of the prediction. In this paper, we seek to significantly improve on both, the method used to generate the explanations and the nature of explanations produced. We use a Generative Adversarial Network\cite{b3} for synthetic data generation and train a piecewise linear model\cite{b14} in the form of Linear Model Trees to be used as the surrogate model. The former enables us to generate a synthetic dataset which better represents the locality of the test point and the latter acts as a stronger and more interpretable surrogate model to capture that locality. In addition to individual feature attributions, we also provide an accompanying context to our explanations by leveraging our surrogate model's structure and property. 
\end{abstract}


\maketitle

\section{Introduction}
The Finance domain has a number of scenarios where sophisticated machine learning models are used to predict various quantities of interest. This ranges from prediction of credit and fraud risk of transactions to digital marketing. In many of these instances, it is not enough to provide accurate predictions, but is also important to provide an intuitive reason in coming up with the prediction that can be understood by a human. In instances like credit approval, it is also mandated by law to provide an explanation to consumer for the decision made for approval or denial of the credit\cite{b19,fairlending}. In applications like credit line extension\cite{b20}, predictions from these sophisticated machine learnt models could be used by humans to arrive at a decision. In this scenario, it is desirable for the human decision maker to understand the reasons behind a particular prediction made by the machine. In all of these applications, it is often the case that the model is used by an individual/team different from the ones who built it. Hence, we require tools to provide \emph{explanations} for the various predictions made by the sophisticated model without relying on the model's internal structure or the builders of the model. In recent times, there have been a number of methods proposed to provide explanations of predictions made by complex models\cite{b18}\cite{b1}\cite{b2}\cite{b16}. Providing an explanation of a complicated model that is applicable globally across the input space is a difficult task. Hence, most of these methods zoom into a locality and explain the local decision boundaries using simple models to explain the decisions made on a test point\cite{b13}. The simple model essentially tries to replicate the behaviour of the complex model (hereafter referred to as \emph{target model}) through simplistic rules or feature attributions. This technique has two steps. First, we generate points around the locality of the test point. This data does not necessarily exist in the training dataset of the user and is therefore synthetic. The next step is to fit another model, called a surrogate model, on this generated data. This model is usually a simple interpretable model like a linear weighted regression model or a set of decision rules. The co-efficients of a linear regression model represents how each feature of the test point contributed to the final prediction and in case of decision rules, it would give the set of rules that a test point followed in order to be classified with a certain label. A surrogate model and its generation process has the following desiderata-
\begin{enumerate}
\item The surrogate model used should be intuitive so that a prediction can be explained using simple feature attributions or decision rules. \label{intuitive}
\item The surrogate model should be accurate in matching the prediction made by the target model that we are trying to explain. \label{powerful}
\item The data generation process used to sample new points should be able to reproduce the data distribution around the test point faithfully.
\end{enumerate} 
In this paper, we propose a methodology for explaining predictions of a sophisticated target model that meets the desiderata listed above with a competitive performance when compared to the state of the art explainability methods. We achieve this using a Linear Model Tree(LMT)\cite{b11} as surrogate. An LMT is a decision tree with linear regression models at each of the leaves. We also use a more precise and powerful sampling method that uses Generative Adversarial Network (GAN) \cite{b3} to generate the synthetic data around the test point. The GAN is built to generate high quality tabular data by considering categorical and numerical data separately. It tries to learn the distribution around the locality using a conditional generator which captures the joint distribution of the variables reliably compared to other sampling methods. Therefore, it is able to provide better local synthetic data for the surrogate model to be later trained on. We call our methodology as Linear Model Tree based Explanation(LMTE).\\
In the desiderata provided above, intuitiveness and accuracy of explanations are often at odds. In our paper, we choose linear model trees that provide a good balance of the two. The non-linearity of LMT makes it powerful enough to model local non-linear decision regions, at the same time retaining the intuitiveness of a decision tree.  By virtue of having the structure of decision trees, LMT is able to provide rules explaining the region of input space around the test point of interest. LMT is also able to provide attributions corresponding to each feature since it has a linear regression model at each leaf. In this way, an LMT surrogate model is able to combine the strengths of rule based explanations with feature attributions based explanations. \\
We start our analysis with a qualitative comparison of our method with the state of the art in Section \ref{sec:related}. In Section \ref{sec:methodology} we discuss the various components of our methodology. Following that, in Section \ref{sec:lmte}, we illustrate some sample explanations that our method provides. Section \ref{sec:eval} introduces and explains the evaluation metrics and comparisons with existing frameworks. 

\begin{table}
\caption{Enumeration of the features of different local explanation methodologies.}
\tiny
 \begin{tabularx}{\columnwidth}{|X|X|X|X|X|X|} 
\hline
Features&LIME&SHAP&ANCHORS&LORE&LMTE\\ 
\hline
Surrogate model&linear regression&generalised linear regression&set of rules&decision tree&linear model trees\\ 
\hline
Sampling method&uniform random with independence assumption&sampled from marginal distribution&bandit pull sampling&genetic algorithm&CTGAN\\
\hline
Form of explanation&feature attributions&feature attributions&rules&rules&feature attributions and rules \\
\hline
Supports Regression Tasks&Yes &Yes&No&No&Yes\\
\hline
\end{tabularx}
\label{tab:compare}
\end{table}

\section{Related work}
\label{sec:related}
With the popularity of sophisticated yet non-interpretable models like neural networks, there have been different methodologies proposed to explain the predictions of these complex models. One of the earliest attempts at this has been LIME\cite{b1}, which uses linear regression(logistic regression for classification) model as a surrogate. This makes feature attributions simple to compute and intuitive to understand. For generating synthetic data around the test point, LIME uses random uniform sampling assuming independence amongst features.  Once the synthetic data is generated, LIME employs weighing schemes to decide the importance of each generated synthetic point. After this, a linearly weighted model is fit to this synthetic data. The goal is to predict labels as close to the labels that the target model would have predicted for the synthetic points. The labels from the target models are obtained by making prediction calls to it. But often, these simple weighted linear models, although interpretable, are not powerful enough to model the synthetic data generated around the locality when the locality is non-linear\cite{b2}. Failure to model the joint distribution of features while sampling, further degrades the ability of the surrogate model  to mimic the distribution of the data around the test point of interest. Moreover, these explanations come with no general guidelines of when they would be applicable beyond the given test point. This lack of context provided with the explanations limits their usage.\\
SHAP\cite{b18} is another method of explaining prediction that is an approximation of Shapley values\cite{b21}. SHAP implicitly assumes a surrogate, which is a generalized linear model. The model's coefficients correspond to the contributions made by the individual features to the prediction made by the model. We could consider SHAP as a generalization of LIME. Like LIME, SHAP also provides the contributions of individual features in making the prediction for a particular test point and hence is highly local in its explanations. \\
The disadvantage of hyper-local explanations which do not consider the context and produce explanations for a single test point is that these explanations would be valid for the specific test point under consideration but can drastically change for a test point in the neighbourhood of the original test point\cite{volatile}. This is counter-intuitive to our expectation. We would like to have explanations that are consistent across a local region around a test point than just specifically for the test point. ANCHORS\cite{b2} overcomes this problem by using decision rules instead of individual feature attributions as explanations. These rules are designed to have high precision, applying to points within a region of space in such a way that the points satisfying the rule would receive the same label with high probability. However, ANCHORS' reliance on precise rules is often at the expense of coverage. This would often result in test points not having a rule covering it. Having rules as explanations works well for classification where we have labels with no specific ordering. However, adapting them to regression problems is often very difficult and sub-optimal. Especially, it would be difficult to comprehend the contribution of features to the magnitude of the prediction made by a regressor, using a set of rules. In these cases, explanation in the form of feature attributions works well.\\
A recent method that improves upon LIME and ANCHORS is LORE\cite{b16}, which uses a decision tree as a surrogate model and a genetic algorithm to sample the space around a test point. The use of a decision tree provides a context for a prediction like ANCHORS. However, it would be still difficult to quantify the attributions of different features to the prediction since the explanations are in the form of decision rules. The other disadvantage of LORE is the use of a genetic algorithm for sampling. The algorithm is not very efficient in sampling, often producing samples that are not reflective of the true distribution in the region of interest. Like ANCHORS, LORE also does not support regression tasks.\\
In our paper, we overcome the problems with the current methods of local explanations by using a combination of an efficient sampling methodology and a powerful model like LMT. LMT as a surrogate enables us to produce rule based explanations to illustrate the context around the test point like ANCHORS and LORE, and the linear regression models in the leaf nodes help us produce feature attributions like LIME and SHAP.  LMT also allows us to support both classification and regression tasks. An LMT model helps us model non-linear decision boundaries producing more accurate explanations without compromising the interpretability of the explanations itself. An efficient sampling method like CTGAN \cite{b4} helps mimic the local regions faithfully, thus producing more contextual LMT models. Table \ref{tab:compare} summarises the features of different methodologies that we compare in this paper.

\section{Methodology}
\label{sec:methodology}

In this section, we provide the methodology used by LMTE to provide explanations of predictions. We provide a detailed analysis of the two components - the data sampling process for training the surrogate model and the surrogate model itself.

\subsection{Generating Local Synthetic Data Around Test Point}\label{sec:datagen}

The rise of Generative Adversarial Network (GAN) has led to new and more robust ways of generating synthetic data in recent times. In a traditional GAN, two neural networks, a discriminator and a generator, compete against each other. The generator tries to create new synthetic data to fool the discriminator, whereas the discriminator attempts to distinguish between real and synthetic data. \\
However, using GAN to generate tabular data is non-trivial. Mode collapses, mixed data types, and class imbalances are some of the issues faced when using GAN for tabular data. But by using new methods such as Conditional Tabular Generative Adversarial Networks (CTGAN) \cite{b4}, we can generate high quality synthetic data. CTGAN, a variant of GAN, is built especially for tabular data and can handle all the problems mentioned above. Recent research\cite{b5} also shows that using transformations such as Box-Cox transformations\cite{b10} for numerical data and one-hot encoding for categorical data leads to a better learning procedure. By incorporating all of these in our method, we generate high quality synthetic data around the locality of the test point.\\
Upon getting a test point, we first find its K nearest neighbours from the training dataset. Numerical columns are then scaled using MinMax Scaling if the values are not positive. The categorical columns are transformed into one-hot vectors. Next, box cox transformation is applied to numerical columns to ensure that the features are distributed normally and do not have long tails. These scaling and encoding transformations are optional and may or may not be used by the user depending on the dataset. This transformed tabular data is then fed as input to the CTGAN along with specifying the discrete columns since this allows CTGAN to handle them separately. After the CTGAN is trained for some epochs, test points can be sampled from the trained model. This set of sampled points is our data generated around the test point’s locality. Finally, since all of the transformations done before were reversible, inverse transformations are applied to the data generated by the CTGAN. This provides synthetic data in the same format as the training data on which the target model was trained. Labels are obtained for this data by making prediction calls to the target model. These are considered as the true labels for this dataset since we are trying to mimic the target model.  We call this local synthetic dataset as T\textsubscript{syn}.

%
%

\subsection{Linear Model Trees As Surrogates}
Piecewise Linear Models are capable of capturing nonlinear patterns without compromising on the interpretability of the model. Each region has a linear model associated with it. Using this, the model can easily be explained by providing the region's context and the coefficient of the linear model associated with it. We use a powerful member of the family of piecewise linear models known as Linear Model Trees\cite{b11}.\\
These are decision trees built by recursively splitting on features that minimize the loss of an associated linear/logistic model. There are both greedy and adaptive search strategies for finding the best split. The greedy strategy is computationally expensive with a large number of features. Therefore, we can use the adaptive search strategy when the feature space is high dimensional. In a greedy strategy, the node is split on each possible feature value for each feature and then the best split is chosen. However, in an adaptive search strategy, for each feature, only a certain number of feature values are candidates for best split, and each value is not tried. This speeds up the process considerably. Each leaf node has a linear model that is fit on it and is a property of the node. These models are much more powerful than simple linear models and also provide greater interpretability. They can capture non linearities and also provide the context of the explanations instead of just the coefficients given by linear models. The prediction of a test point is made using the linear model of the leaf that it falls into when it traverses the tree. \\
We provide a twofold explanation of the test point. The first is the context of the explanation. This is the path that the test point traversed on its way to the leaf node. This can be provided as a set of rules consisting of the feature and feature values using the thresholds set at the nodes. The second is the coefficient of the linear model that was fit on the leaf node in which the test point fell. By providing both the context and the coefficient, we provide a comprehensive picture of the prediction. The decision path taken by the test point provides information about the input region where the model applies and the coefficients of the linear model provide the attributions to the individual features. It must be noted that this in contrast with other methods were we can get only  one of these.

\subsection{Linking Data Generation And Surrogate Modelling }\label{AA}
We will now briefly illustrate the end to end working of our post hoc, model agnostic explanation methodology. Our inputs are the original training data, T\textsubscript{org}, the trained complex target model, $f(x)$, and a test point $X\textsubscript{t}$. Our goal is to explain the prediction of $X\textsubscript{t}$ from $f(x)$, called $Y\textsubscript{t}$ . We take the test point $X\textsubscript{t} $ and get its $K$ nearest neighbours from $T\textsubscript{org}$. After this, the transformations discussed in Section \ref{sec:datagen} are performed and the synthetic data, $T\textsubscript{syn}$  is obtained from the CTGAN. Next, the labels $Y\textsubscript{syn}$ are obtained for this by making calls to $f(x)$. Having both $T\textsubscript{syn}$ and $Y\textsubscript{syn}$, we are now ready to model this locality around the test point using a surrogate model. For classification tasks, Linear Model Trees fitted with Logistic Regression models at nodes are used. For regression tasks, there are Linear Regression models at nodes instead.  Once this Linear Model Tree is trained on $T\textsubscript{syn}$ with $Y\textsubscript{syn}$ as labels, our test point is passed into this surrogate model. Twofold explanations for the prediction are of the test point are eventually produced. This end to end methodology is agnostic to the target model and the Linear Model Trees are capable of generating these twofold explanations irrespective of whether the target model is tree-based.\\
\begin{figure}[ht]
\makebox[\linewidth]{
  \includegraphics[keepaspectratio=true,scale=0.4]{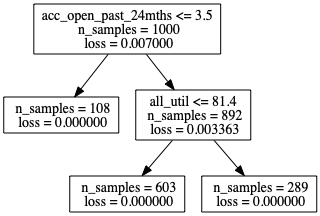}}
  \caption{Surrogate Tree Structure for explanation.}
  \Description{What to write here?}
  \label{vis5}
\end{figure}

\begin{figure}[ht]
\begin{subfigure}{0.5\textwidth}
\includegraphics[keepaspectratio=true,scale=0.25]{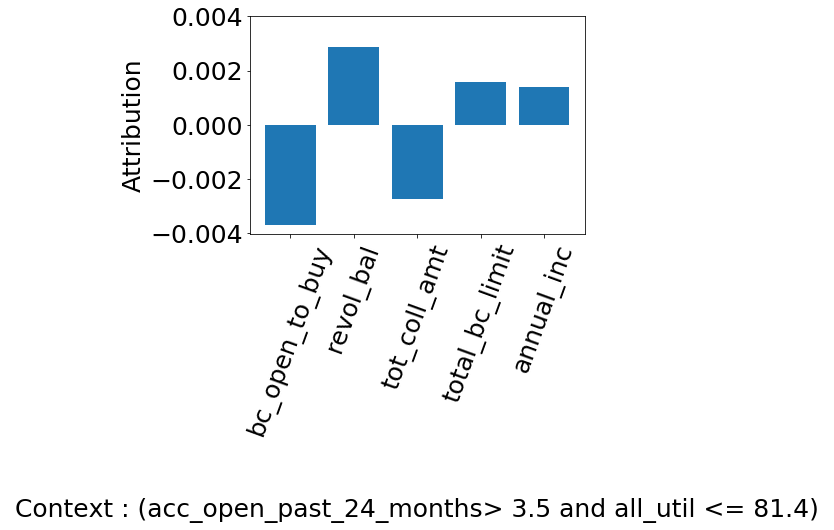}
  \caption{Top 5 most important features that explain the prediction of the original test point.}
  \label{vis6}
\end{subfigure}
\begin{subfigure}{0.5\textwidth}
 \includegraphics[keepaspectratio=true,scale=0.25]{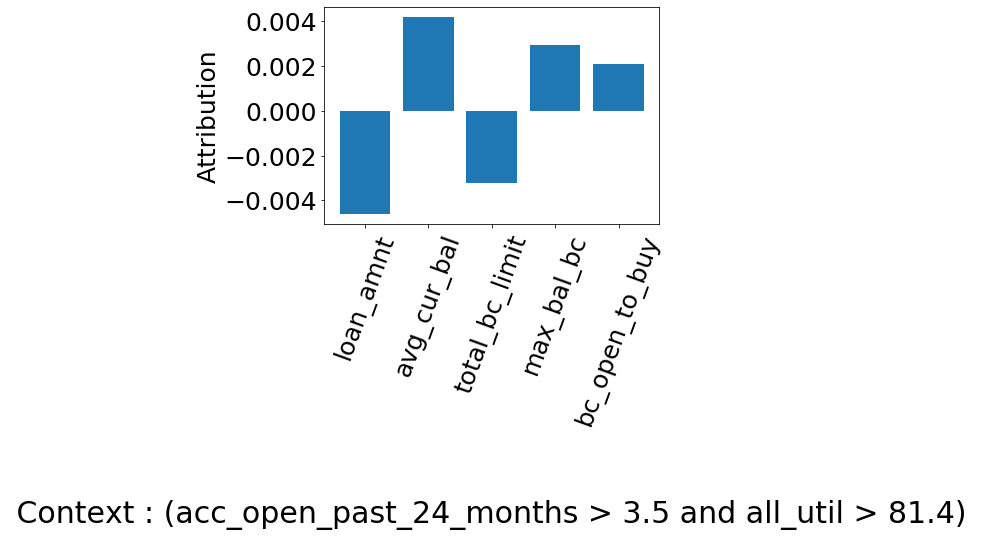}
  \caption{Top 5 most important features that explain the predictions of the hypothetical test point with the value of  "all\textunderscore util" feature being different.}
  \label{vis7}
 \end{subfigure}
 \caption{Feature attributions obtained from the LMTE model for the original test point and a "what-if" condition.}
 \label{fig:feat}
\end{figure}

\section{LMTE explanations}
\label{sec:lmte}

To illustrate some examples of our framework's predictions, we take a public dataset that has data from LendingClub.com\cite{b15}  for the year of 2007. This dataset assesses a loan lending decision problem. We pre-process this data and use 10,000 samples with 75 numerical features to train a Random Forest Classifier, which we seek to explain. In this example, we provide explanations for a test point which has features "acc\textunderscore open\textunderscore for\textunderscore past\textunderscore 24mths" equal to 24 and "all\textunderscore util" equal to 78. We are not listing the values of other features here due to lack of space. Using our LMTE framework, we obtain clear and intuitive explanations of the trained target model's prediction for this test point. The context for this test point is set by the tree structure, as shown in \figurename{~\ref{vis5}}. Specifically, the context is that  "acc\textunderscore open\textunderscore for\textunderscore past\textunderscore24mths"$\geq$3.5 and "all\textunderscore util"$\leq$81.4. The top 5 feature attributions for this prediction can be seen in \figurename{~\ref{vis6}} with the context below satisfying this test point.\\
The learnt tree structure in \figurename{~\ref{vis5}} also allows us to get attributions for "what-if" conditions. By looking at the linear models corresponding to other contexts, we can get a sense of the feature attribution in such cases. This allows us to generate explanations for "what-if" conditions. For example, if our test point had everything else same except the "all\textunderscore util" value, which we now set to some value greater than 81.4, we would have attributions as shown in \figurename{~\ref{vis7}}. Notably, the top features that affect the model predictions in both these cases are different. The tree structure allows us to enclose our explanations in the context in which these attributions are valid, and the linear model corresponding to that context gives us the feature attributions.

\begin{table*}
  \caption{Performance of the 3 surrogate models on synthetic data generated by LIME in different tasks.}
  \label{tab:fidelity}
\scalebox{0.9}{\begin{tabular}{|l|l|l|l|l|l|l|l|} 
\hline
Surrogate Model & \multicolumn{7}{l|}{~ ~ ~ ~ ~ ~ ~ ~ ~ ~ ~ ~ ~ ~ ~ ~ ~ ~ ~ ~ ~ ~ ~ ~ ~ ~ ~ ~ Dataset \textbar{} Task}  \\ 
\hline
                & ForestFire (R) & Abalone (R) & AutoMpg (R) & Hearts (C) & Breast (C) & BankNote(C) & PIMA (C)         \\ 
\hline
LIME            & 0.83          & 0.46        & 0.58        & 0.89       & 0.93       & 0.88        & 0.90             \\ 
\hline
Decision Tree Classifier (LORE) & -          & -        & -       & 0.90       & 0.98       & 0.99        & 0.97             \\
\hline
Linear Model Tree & \bf{0.56}          & \bf{0.33}        & \bf{0.40}       & \bf{0.98}       & 0.98       & 0.99        & 0.97             \\
\hline
\end{tabular}}

\end{table*}

\begin{table*}
  \caption{Fidelity Scores when methods are trained on their own neighbourhood and tested on both their neighbourhood and that of the opposing method.}
  \label{tab:crisscross}
\scalebox{0.8}{\begin{tabular}{|l|l|l|l|l|l|l|l|l|l|l|l|l|l|l|} 
\hline
                  & \multicolumn{14}{l|}{Dataset \textbar{} Task}                                                                                                                                                                                                                     \\ 
\hline
                  & \multicolumn{2}{l|}{ForestFire (R)}  & \multicolumn{2}{l|}{Abalone (R)} & \multicolumn{2}{l|}{AutoMpg (R)}            & \multicolumn{2}{l|}{Hearts (C)} & \multicolumn{2}{l|}{Breast (C)~} & \multicolumn{2}{l|}{BankNote (C)~} & \multicolumn{2}{l|}{Pima (C)~}  \\ 
\hline
Surrogate Model   & \multicolumn{2}{l|}{~ ~ ~ ~ ~ ~ ~ ~} & \multicolumn{2}{l|}{}            & \multicolumn{2}{l|}{Data Generation Method} & \multicolumn{2}{l|}{}           & \multicolumn{2}{l|}{}            & \multicolumn{2}{l|}{}              & \multicolumn{2}{l|}{}           \\ 
\hline
                  & CTGAN & URS                          & CTGAN & URS                      & CTGAN & URS                                 & CTGAN & URS                     & CTGAN & URS                      & CTGAN & URS                        & CTGAN & URS                     \\ 
\hline
Linear Model Tree & 0.55  & 14.29                      & 0.42  & 1.70                     & 0.24  & 1.20                               & 0.98  & 0.74                    & 0.99  & 0.62                     & 0.99  & 0.81                       & 0.98  & 0.79                    \\ 
\hline
LIME              & 1.17 & 0.83                        & 1.71  & 0.46                     & 0.89  & 0.58                                & 0.72  & 0.89                    & 0.91  & 0.93                     & 0.97  & 0.88                       & 0.83  & 0.90                    \\
\hline
\end{tabular}}

\end{table*}
\section{Evaluation}
\label{sec:eval}
To compare our LMTE method with existing methods, we use both artificial and real-world datasets. We begin by visually inspecting the neighbourhood generated by the different methods and then proceed to several quantitative measures of comparison between our method and the current state of the art. In all of the comparisons provided in this section, we focus on LIME, ANCHORS, and LORE. We do not include SHAP in our quantitative evaluations since the surrogate models and the generative distributions in SHAP is difficult to separate as SHAP implicitly creates these.

\subsection{Artificial Data}
\begin{figure}[ht]
\begin{subfigure}{0.5\textwidth}
\includegraphics[keepaspectratio=true,width=0.8\textwidth]{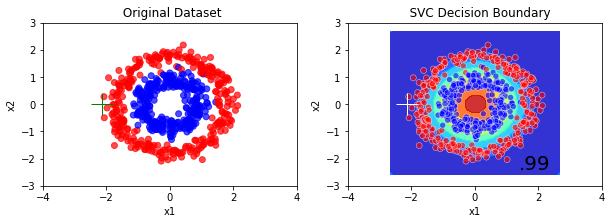}
  \caption{The original dataset(left) and the decision boundary of a Support Vector Classifier (SVC) trained on it.}
  \label{vis1}
\end{subfigure}

\begin{subfigure}{0.5\textwidth}
\includegraphics[keepaspectratio=true,width=0.8\textwidth]{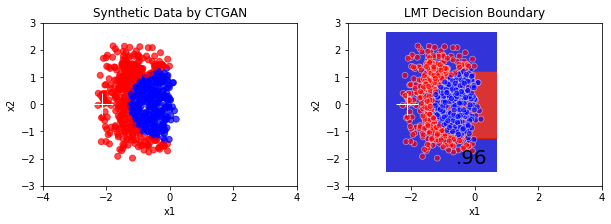}
  \caption{Synthetic dataset generated by CTGAN(left) and the decision boundary of a Linear Model Tree. }
  \label{vis2}
\end{subfigure}

\begin{subfigure}{0.5\textwidth}
 \includegraphics[keepaspectratio=true,width=0.8\textwidth]{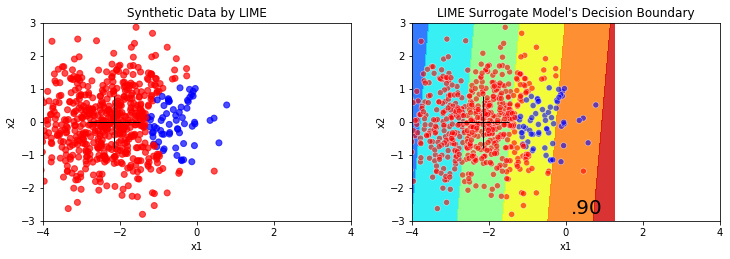}
  \caption{Synthetic dataset generated by LIME(left) along with the decision boundary of the weighted Linear Regression model}
  \label{vis3}
 \end{subfigure}
 \begin{subfigure}{0.5\textwidth}
 \includegraphics[keepaspectratio=true,scale=0.1,width=0.8\textwidth]{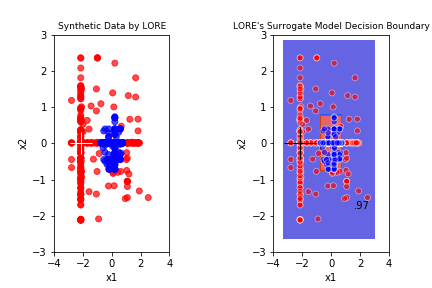}
  \caption{Synthetic dataset generated by LORE(left) along with the decision boundary of the Decision Tree Classifier.}
  \label{vis4}
 \end{subfigure}

 \caption{Comparison of the synthetic data generated by various methods and the corresponding decision boundaries learnt by the corresponding surrogate models after training on their respective synthetic data. The candidate test point is represented by a plus sign in each of these visualisations.The number on bottom right represents the fidelity of surrogate model with respect to SVC on its synthetic data. The color scheme of the decision boundaries is blue for when the model is confident of class being red and vice versa. Greenish yellow spectrum shows uncertainty in prediction}
 \label{fig:feat}
\end{figure}

We use a 2-dimensional artificial dataset to illustrate the differences between current methods and LMTE visually. The training dataset we are trying to model has two features $x1$, and $x2$ and two classes, blue and red and is plotted in \figurename{\ref{vis1}}. A target machine learning model in the form of a Support Vector Classifier (SVC) is trained on this dataset. The decision boundaries of the SVC model is also shown in \figurename{\ref{vis1}}. This SVC fits the training data well, which is shown by an accuracy score of 0.99 on the data. \\
Now, let us visually see what kind of synthetic data is being generated by our method. Given the test point marked with a plus sign, using our method, we generate the synthetic data that closely represents the true data distribution of the training samples, as seen in \figurename{\ref{vis2}}. On the other hand, data generated by LIME in \figurename{\ref{vis3}} and the Genetic Algorithm method in LORE in \figurename{\ref{vis4}} has considerable noise and is scattered which is not a good reflection of the local neighbourhood around the test point. \\
We can also see how well our surrogate models can model the synthetic data which are generated by their respective methods. As seen in \figurename{\ref{vis2}}, Linear Model Trees are strong surrogates that can capture non-linearities and fit the synthetic data very well. The labels produced by this surrogate agrees with the labels obtained from the trained SVC on this data with an accuracy score of 0.96. The weighted Linear Regression model which is used as a surrogate in LIME, however, does not do a very good job of fitting the synthetic data generated for itself, as seen in \figurename{\ref{vis3}}. The region with a greenish-yellow spectrum in the middle indicates that the surrogate model is unable to distinguish between the points in this region. The linear model gets an accuracy score of 0.90 on the synthetic data. The simplistic linear boundary is unable to deal with the non-linear boundaries in the neighbourhood.

\begin{table*}
  \caption{Comparison of LIME, LORE and LMTE on their fidelity.}
  \label{tab:e2e}
\scalebox{0.9}{\begin{tabular}{|l|l|l|l|} 
\hline
     & German & Compass & Adult  \\ 
\hline
LIME & 0.87   & 0.88    & 0.87   \\ 
\hline
LORE & 0.99   & 0.99    & 0.99   \\ 
\hline
LMTE & 0.97   & 0.95    & 0.95   \\
\hline
\end{tabular}}

\end{table*}

\begin{table*}[ht]
  \caption{Coverage and Precision of rules generated by LORE, Anchor and LMTE.}
  \label{tab:anchors}
\scalebox{0.9}{\begin{tabular}{|l|l|l|l|l|l|l|} 
\hline
       & \multicolumn{2}{l|}{German} & \multicolumn{2}{l|}{Adult} & \multicolumn{2}{l|}{Compass}  \\ 
\hline
       & Coverage & Precision        & Coverage & Precision       & Coverage & Precision          \\ 
\hline
ANCHORS & 0.20     & 0.99             & 0.075    & 0.97            & 0.025    & 0.90               \\ 
\hline
LORE   & 0.20     & 0.96             & 0.25     & 0.89            & 0.48     & 0.13               \\ 
\hline
LMTE   & 0.48     & 0.79             & 0.64     & 0.80            & 0.45     & 0.76               \\
\hline
\end{tabular}}

\end{table*}

\subsection{Comparison of Surrogate Models}

The fidelity of the surrogate model with the target model is used to measure how well the surrogate is able to mimic the target model. For classification tasks, the surrogate model's fidelity is defined as the proportion of instances on which the labels generated by the original target model are the same as the predictions produced by the surrogate model. The higher the proportion of these instances, the better the surrogate model is. For regression tasks, the fidelity is defined as the Root Mean Squared Error between the real-valued predictions generated by the target model and the surrogate model on the synthetic locality. We standardize this RMSE by dividing with the standard deviation of the true labels, which are the  predictions given by the target model. The lower the value of this standardized RMSE, the better the surrogate model is. \\
We have 3 surrogate model candidates. The weighted Linear Regression from LIME, Decision Tree Classifier from LORE, and Linear Model Trees from LMTE. ANCHORS is not included in this comparison because it does not use this approach of training a surrogate model and instead performs a search for rules. To compare the modelling power of these surrogate models, we train and test them on the same dataset. We choose the neighbourhood generated by LIME to be this dataset(Table \ref{tab:fidelity}).\\
This is done for 7 public datasets from the UCI repository\cite{b6}. There are 4 classification tasks and 3 regression tasks. Since LORE uses a decision tree classifier, we were unable to run it for the 3 regression tasks. For all tasks, the following setup is used to ensure a fair comparison. The complex target model which is being explained is a Random Forest in all cases. The dataset is split into train and test. The test split has 25 points while the remaining are put into train. With the Random Forest model being trained on the training data, the methods are evaluated on a total of 25 test points. For each run with a test point, both data generation methods are made to produce 500 synthetic points. We set the number of nearest neighbours to be 20 in all cases for LMTE. For LIME, we use their open-source implementation. In their API, we set sampling from around the point to be True and discretizing continuous features to be False to keep the sampling settings similar. Using LIME's API, we are able to get the data that it generated for the explanation. However, this data is scaled. So, we inverse transform it to allow it to be tested by our models. In classification tasks, our linear model tree has a maximum depth of 4 with an adaptive search of 50 points. A logistic regression model with L2 penalty is fit in the nodes. This is also called as a Logistic Model Tree \cite{b7}. In regression tasks, we use maximum depth of 2 for linear model trees with a linear regression model at the nodes. For LORE, the default parameters set in their scripts are used. The mean of fidelity scores for 25 test points are reported in classification tasks. In regression tasks, the median of standardized RMSE fidelity values are reported for 25 test points.\\
The non-linear surrogate models of both LMTE and LORE outperform the weighted linear regression model used by LIME. In regression tasks, LMTE is the only alternative to LIME since LORE is only built for classification. On classification tasks, LMTE matches the performance of LORE, outperforming LORE on one of the datasets. The results highlight the benefits of a non-linear surrogate model over a linear one along with the flexibility of having a linear model tree in performing both regression and classification with equal ease and accuracy.

\subsection{Generalization}

Another desirable property is that the surrogate model generalizes well in the neighbourhood. So instead of training and testing on the same neighbourhood, we also test on a different neighbourhood for each method. The Linear Model Tree surrogate is trained on the synthetic data generated by CTGAN and evaluated on that training data and the data generated by LIME for the particular test point. The same is done with the surrogate of LIME. The weighted Linear Regression model is trained on synthetic data generated by LIME using uniform random sampling (URS) and is tested on this and also the data which was generated by CTGAN for the test point(Table \ref{tab:crisscross}). \\
The weighted Linear Regression model used in LIME performs better on the held-out synthetic data generated by CTGAN than Linear Model Trees do on held out data generated by LIME. This shows that the data generated by CTGAN represents the locality better with lesser noise as compared to data generated by URS. The added noise makes it difficult for the Linear Model Tree to perform well on it since it was modelled on the locality of the test point according to the distribution learned by CTGAN which was not noisy.

\subsection{Comparison of the combined system}

In this section, we compare the methodologies of all the explanation systems in its entirety. We use 3 tabular public datasets. The Compass dataset from ProPublica\cite{b17} along with Adult and German dataset from UCI repository\cite{b6}. All datasets have both continuous and categorical attributes. The experimental setup and pre-processing is similar to the experimental setup used in LORE\cite{b16}. We run the 3 frameworks : LIME, LORE, and LMTE, end to end on these 3 datasets. The target model is a Random Forest Classifier with 100 estimators. All data generation methods in the respective frameworks create a local neighbourhood of 500 points. Fidelity is calculated for 50 test points in each of the datasets, and the mean values are reported.(Table \ref{tab:e2e}). The results show that both LORE and LMTE outperform LIME and have very high fidelities.

\begin{table*}
  \caption{Recall Scores.}
  \label{tab:faithfulness}
\scalebox{0.9}{\begin{tabular}{|l|l|l|l|} 
\hline
        & Breast & Heart & Pima  \\ 

\hline
Lime    & 0.72   & 0.81  & 0.98  \\ 
\hline
LMTE & 0.75   & 0.82  & 0.82  \\
\hline
\end{tabular}}

\end{table*}

\subsection{Quality of Contexts}

To measure how good the context provided in our explanations is, we compare it with the rules given by ANCHORS\cite{b2} and LORE\cite{b16}. We use two metrics for the comparison. First is coverage, which is the proportion of samples that obey the rule given in the explanation. The second metric is precision. This is the proportion of covered samples that have the same prediction as the target model. In case of ANCHORS and LORE, the prediction for all covered points is the same class as that of the test point. For our method, we use the surrogate linear model tree to get the local class prediction. All three are matched with the prediction of the target model as the true label. 
We make this comparison on 3 public datasets and report the results (Table \ref{tab:anchors}). We split the dataset into train and test. The coverage and precision numbers are reported using the median value obtained for the first 200 test points in batches of 100 each. The mean of the two runs in reported. A Random Forest Classifier with 20 estimators was chosen as the target model. There is a coverage and precision tradeoff between the methods. Though ANCHORS have high precision, they cover very few points, and their rules are also generally very specific. On the other hand, LMTE has higher coverage implying more generalized rules and has a lower precision. The rules given by LORE lie in the middle.

\subsection{Faithfulness To Model}

We measure the faithfulness of the explanations to the original model. To do so, an inherently interpretable model such as a decision tree of low depth is trained, and its explanation is compared with the explanation given by post hoc methods. In the case of a decision tree, we take the path traversed by the point in the tree to be the explanation. We take the features that were in the path to be the true explanation. To see how faithful the explanations generated by post hoc methods are, we see how many features in the top 50 percent based  on attribution values overlap with the features in the true explanation. This is the recall of the post hoc explanation. 
We do this for 3 public datasets and compare our method with LIME. The results (Table \ref{tab:faithfulness}) show that both methods maintain high levels of faithfulness across datasets. The post hoc explanations are able to achieve a high recall indicating that they can capture similar features as the original model and attribute high importance to them.

\section{Conclusions And Future Work}
In this paper, we proposed new ways to enhance the explanations produced by post hoc model agnostic methods. We introduced LMTE, a framework that generates better synthetic data around a test point's locality using CTGAN and models that locality using more powerful and interpretable surrogates in the form of Linear Model Trees. This method provides more intuitive interpretations with the aid of rules in addition to values corresponding to each feature. This method performs well in comparison to the current state of the art methods on both aspects of generating the data representing localities and in modelling them. 

Going ahead, we would like to make this method applicable beyond tabular data to other forms of data, such as image and text. We would also like to work on theoretical guarantees on aspects like choosing the number of optimal $k$ nearest neighbours required and finding better ways to set hyperparameters such as the number of epochs for the CTGAN and depth of Linear Model Trees. Besides that, work can be done on speeding up the training of the Linear Model Tree as it takes considerable time when there are a large number of features.

\bibliographystyle{ACM-Reference-Format}
\bibliography{kdd_LMTE}

\end{document}